\documentclass[runningheads]{llncs}

\usepackage{eccv}

\usepackage{eccvabbrv}

\usepackage{graphicx}
\usepackage{booktabs}
\usepackage{multirow}
\usepackage{tikz}
\usepackage{orcidlink}
\usepackage{flafter}
\usetikzlibrary{arrows.meta,positioning,fit,calc}

\usepackage{hyperref}
\hypersetup{
  pdftitle={GuideFetch: A Task Coordination Framework for Concurrent Navigation and Object Retrieval in Assistive Robot Dogs}
}

\graphicspath{{figures/}}

\newcommand{\method}{\textsc{GuideFetch}}

\newcommand{\taskgraph}{\mathcal{G}}
\newcommand{\robots}{\mathcal{R}}
\newcommand{\capset}{\mathcal{C}}
\newcommand{\actions}{\mathcal{A}}

\makeatletter
\renewcommand*{\@fnsymbol}[1]{\ensuremath{\ifcase#1\or *\or \dagger\or \ddagger\or
    \mathsection\or \mathparagraph\or \|\or **\or \dagger\dagger
    \or \ddagger\ddagger \else\@ctrerr\fi}}
\makeatother

\begin{document}

\title{GuideFetch: A Task Coordination Framework for Concurrent Navigation and Object Retrieval in Assistive Robot Dogs}
\titlerunning{GuideFetch: Coordinated Navigation and Retrieval}
\author{
Qian Yin\inst{1}\orcidlink{0009-0005-4540-5313} \and
Ruiping Liu\inst{1,}\thanks{\textbf{Corresponding authors:} ruiping.liu@kit.edu; kunyu.peng@kit.edu}\orcidlink{0000-0001-5245-2277} \and
Kunyu Peng\inst{1,*}\orcidlink{0000-0002-5419-9292}\and
Jianxiang Man\inst{1}\and
Isik Baran Sandan\inst{1}\and
Junwei Zheng\inst{1}\orcidlink{0009-0005-4390-3044}\and
Yufan Chen\inst{1}\orcidlink{0009-0008-3670-4567} \and
Di Wen\inst{1}\orcidlink{0009-0000-1693-7912}\and
Kailun Yang\inst{2}\orcidlink{0000-0002-1090-667X}\and
Rainer Stiefelhagen\inst{1}\orcidlink{0000-0001-8046-4945}
}

\authorrunning{Qian Yin~\textit{et al.}}
\institute{Karlsruhe Institute of Technology, Germany \and
Hunan University, China}
\maketitle
\begin{abstract}
Consider one robot guide dog escorting a blind user to a seat while a second
retrieves and delivers an object. We introduce \textsc{GuideFetch}, a framework
for coordinating this concurrent guide-and-fetch mission with heterogeneous
robots. A large language model (LLM) instantiates a schedule-conditioned
four-action schema; deterministic
normalization and validation enforce registered targets, robot capabilities,
and the selected schedule, while robot and object states govern execution and
completion. We record 360 simulator runs over 90 scene--seed combinations under
scripted and online plan-provenance conditions. All 180 online responses validate on the
first request and match their scripted references, so the plan-provenance comparison
tests normalized-plan agreement rather than a distinct execution factor. A simulator-free mutation
test accepts two valid controls and rejects all 32 rule-violating variants.
Across 90 scene--seed cases per schedule, sequential and parallel execution achieve $72/90$
and $71/90$ operational successes. Among 56 common successes, the implemented
role-reassigned parallel protocol reduces mean makespan by 41.3\%. This
system-level gain combines role assignment, action overlap, and scene geometry;
state checks distinguish plan validity from verified mission completion.
\end{abstract}
\section{Introduction}
\label{sec:introduction}

Robot guide dogs are increasingly expected to support blind and low-vision
users beyond simple point-to-point navigation, including tasks that require
retrieving and delivering everyday objects. Consider one such scenario: when
the user wants to rest, one guide dog leads them to an available seat while a
second assistive robot dog is dispatched to a coffee machine, retrieves a cup, and
delivers it to the same seat. Realizing this behavior requires a robot team to
satisfy navigation and object-retrieval objectives within a single mission,
even though the two capabilities belong to different embodiments. Assistance
dogs already exhibit this kind of functional specialization in practice, with
guide, hearing, and service dogs each trained for a distinct
task~\cite{adi2026definitions}. In our setting, a \emph{guider} completes the
navigation objective while a \emph{fetcher} acquires and delivers the object.
Once these responsibilities are distributed across heterogeneous robots,
however, coordination requires more than executing two independent skill
sequences: each action must be assigned to a capable embodiment, temporal
dependencies must be respected, conflicting commands to the same robot must be
prevented, and mission progress must be verified from measured terminal conditions rather
than assumed from plan validity or motion smoothness alone.

The two capabilities have largely developed separately. Guide-robot research
addresses wayfinding and user safety~\cite{kim2023guide}; mobile-manipulator
research addresses object retrieval~\cite{king2012dusty}. Language-based systems connect requests to
predefined skill affordances~\cite{ahn2023saycan}, executable
programs~\cite{liang2023code,singh2023progprompt}, and execution
feedback~\cite{huang2023inner}. Multi-robot extensions use large language
models (LLMs) for decomposition, dialogue, and capability-aware
allocation~\cite{kannan2024smart,madi2024roco,liu2025coherent,chen2025emos}.
Yet one question remains for guide-and-fetch assistance. Can heterogeneous
robots exploit concurrency while ensuring that symbolic progress corresponds
to verified task completion?

The generated plan alone cannot answer this question. A valid-looking plan may
assign grasping to a robot without a manipulator. A completed trajectory may
leave the object unsupported, while an early failure may make a parallel run
appear fast. The guider must reach its destination, and carrying must remain
blocked until the fetcher reaches pickup, establishes contact, and lifts the
object. We therefore treat plan validity, execution progress, and mission
completion as distinct states.

\method{} separates task specification from robot execution and measured
completion. A schedule-conditioned language interface uses an LLM to instantiate
a schema over registered actions. A deterministic validator resolves aliases and rejects assignments
that violate robot capabilities or the selected schedule. The accepted records
are compiled into nominal dependency levels, and branch-local state machines
execute under one simulation clock. Navigation arrival, bilateral gripper
contact, object lift and retention, and assisted delivery are recorded as
terminal events that govern branch progress and mission success. The LLM
supplies structured action records, while deterministic checks and embodied
state govern execution and completion.

\begin{figure}[t]
\centering
\includegraphics[width=\textwidth]{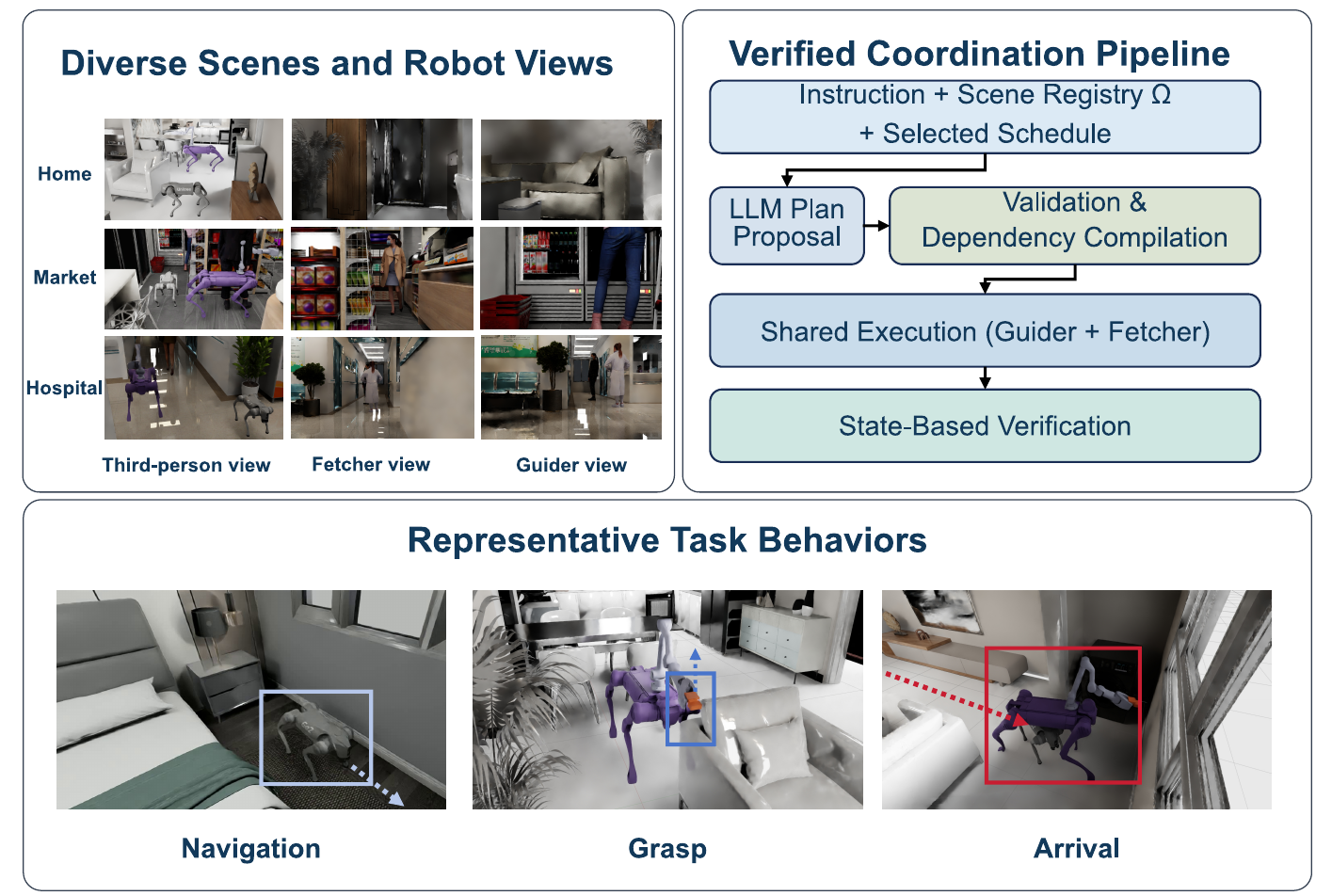}
\caption{Overview of \method{}.  Left: three evaluation scenes and robot
views.  Right: schedule-conditioned plan instantiation, deterministic validation and dependency compilation, shared execution of the guider and fetcher, and state-based
verification.  Bottom: representative navigation, grasp, and arrival
behaviors.}
\label{fig:overview}
\end{figure}

We instantiate \method{} in InternUtopia, built on
GRUtopia~\cite{wang2024grutopia}, with a Unitree Go2 guider and a Unitree B2
fetcher equipped with a Z1 arm. Sequential execution assigns the destination visit, pickup, and
delivery to the fetcher; parallel execution assigns the unloaded visit to the
guider while retrieval begins. Crossing both schedules with scripted and fresh
online plans over 90 matched scene--seed cases yields 360 executions.

All 180 online responses match their scripted references without fallback or
replay, making the plan-provenance comparison a normalized-plan agreement check. A mutation test accepts
two controls and rejects 32/32 rule-violating variants. Sequential and parallel
schedules achieve $72/90$ and $71/90$ successes; over 56 common successes, the
implemented parallel protocol reduces mean makespan from 164.78 to 96.79~s
($41.3\%$). The gain reflects redistributed and overlapping work, while all
failures occur during pre-transport manipulation.
\section{Related Work}
\label{sec:related}

\noindent \textbf{Assistive navigation and object retrieval.}
Robot guidance and robotic object assistance address complementary needs, but they have often been developed as separate systems. ARP~\cite{kim2023guide} formalizes wayfinding, interaction, and safety for a
quadruped guide robot. Dusty~\cite{king2012dusty} demonstrates object retrieval
with an assistive mobile manipulator. More recent navigation work studies
spatial language maps~\cite{huang2025multimodal}, language-guided semantic
graphs~\cite{li2026hosg}, voice interaction with quadruped guide
robots~\cite{qayyum2026toward}, and language and vision cooperation for
assistive mobile robots~\cite{yang2026lavicon}. On the manipulation side,
language has been used to reason about object orientation~\cite{qi2026sofar},
grasp affordances~\cite{chen2026enhancing}, and fine-grained object
affordances~\cite{wang2025affordbot}. We combine registered destinations and
objects with existing controllers to study how a guider and fetcher complete
one request at a shared destination.

\noindent \textbf{From instructions to executable robot behavior.}
Language-based integration requires a user request to become actions that each
robot can execute. SayCan~\cite{ahn2023saycan}
ranks predefined skills using language and learned affordances.  Code as
Policies~\cite{liang2023code} and ProgPrompt~\cite{singh2023progprompt} generate
programs over controlled robot interfaces, while SayPlan~\cite{rana2023sayplan}
grounds longer plans in a three-dimensional (3D) scene representation. Inner Monologue~\cite{huang2023inner} incorporates observations during planning, and recent systems also study verification or repair during execution~\cite{guo2026planar,tripathi2026repair2skill,lima2026agentic}. A well-formed output, however, does not prove that the assignee has the required capability or that an object was lifted and delivered. In \method{}, an LLM instantiates a schedule-conditioned four-action schema, deterministic checks validate its assignments, the compiler constructs its dependency structure, and embodied state determines progress and completion.

\noindent \textbf{Task allocation in heterogeneous robot teams.}
An executable skill set still leaves the team-level decision of which robot
should act and when. Kannan et al.~\cite{kannan2024smart} form coalitions from
robot capabilities, Madi et al.~\cite{madi2024roco} use inter-robot dialogue
and collision feedback, and Liu et al.~\cite{liu2025coherent} integrate
planning and execution for heterogeneous teams. Embodiment-aware work
represents robot capabilities explicitly~\cite{chen2025emos}, while centralized
and decentralized LLM architectures are compared in~\cite{chen2024scalable}.
Related work combines
language reasoning with formation control~\cite{xue2025formation}, cooperative
transport~\cite{xie2025centralized}, and search over joint
plans~\cite{zu2025collaborative}. Our experiment focuses on a narrower question.
When guidance and retrieval belong to different embodiments, can heterogeneous
role assignment and action overlap shorten the mission while preserving its
terminal objectives? We compare sequential and parallel schedules under the
same controllers and state-based terminal checks.

\noindent \textbf{Embodied simulation and controlled evaluation.}
The scheduling comparison requires matched initial conditions and physical
perturbations while preserving the terminal objectives. Habitat~3.0~\cite{puig2024habitat}
supports collaborative tasks involving humans and robots, and
Chang et al.~\cite{chang2025partnr} evaluate planning and reasoning in embodied
multi-agent tasks. GRUtopia~\cite{wang2024grutopia} provides interactive scenes
and configurable tasks for different robots. RoboFactory~\cite{qin2025robofactory}
introduces compositional constraints, while Kang et al.~\cite{kang2025viki}
study embodied multi-agent cooperation. Our study pairs both schedules by scene
and seed while fixing the mission, controllers, and perturbations. The paired
makespan comparison therefore reflects the implemented role assignment and
action overlap among missions completed by both schedules.
\section{Problem Formulation}
\label{sec:formulation}

We formulate the guide-and-fetch mission as a capability-constrained task graph
and compare two schedules that preserve its terminal objectives while changing
robot assignment and action overlap. Let
$\robots=\{r_{\mathrm G},r_{\mathrm F}\}$ denote the guider ($\mathrm G$) and
fetcher ($\mathrm F$), where each robot $r_i$ provides a set $\capset_i$ of
available skills.  The mission vocabulary is
$\actions=\{\texttt{navigate},\texttt{move},\texttt{grasp},\texttt{carry}\}$,
and the scene registry $\Omega$ contains the pickup station $g_p$, final station
$g_f$, and task object $o$. Given an instruction $x$ and a selected schedule,
the language interface instantiates four action records
\begin{equation}
V=\{v_j\}_{j=1}^{4},\qquad v_j=(k_j,r_j,a_j,z_j,p_j),
\label{eq:task_graph}
\end{equation}
where a vertex records the nominal dependency-level index $k_j$, assignee $r_j$, skill
$a_j\in\actions$, target $z_j$, and additional parameters $p_j$. The validator
checks the records against $\Omega$, the capability sets, and the selected
schedule. The compiler then constructs $\taskgraph=(V,E)$, where an edge
$(v_j,v_\ell)\in E$ requires $v_j$ to succeed before $v_\ell$ can start. The
graph is admissible only when every robot and target is registered, each skill
belongs to the assignee's capability set, and the required action structure is
preserved. During execution, every action has one runtime state
\begin{equation}
q_j\in\{\texttt{pending},\texttt{running},\texttt{succeeded},
\texttt{failed}\}.
\label{eq:action_state}
\end{equation}
An action can run only after its predecessors have succeeded and its robot is
idle. Success releases the next action on the same branch. A failed prerequisite
prevents its downstream actions from running and ends the evaluated episode.
The LLM therefore supplies structured action records, while the compiler,
capability checks, and runtime state determine execution.

The sequential and parallel schedules encode the same two objectives, an
unloaded visit to $g_f$ and delivery of $o$ from $g_p$ to $g_f$.  Sequential
execution assigns the complete mission to the fetcher
\begin{equation}
P_{\mathrm{seq}} :=
[\mathrm{F}{:}\mathrm{move}(g_f),
 \mathrm{F}{:}\mathrm{move}(g_p),
 \mathrm{F}{:}\mathrm{grasp}(o),
 \mathrm{F}{:}\mathrm{carry}(o,g_f)].
\label{eq:seq_plan}
\end{equation}
The fetcher first visits the final station without the object, returns to the
pickup station, and then retrieves and delivers the object.  Parallel execution
assigns the unloaded visit to the guider and starts it together with the fetcher
branch
\begin{equation}
P_{\mathrm{par}} :=
\left\{
\mathrm{G}{:}\mathrm{navigate}(g_f)
\mathbin{\parallel}
[\mathrm{F}{:}\mathrm{move}(g_p),
 \mathrm{F}{:}\mathrm{grasp}(o),
 \mathrm{F}{:}\mathrm{carry}(o,g_f)]
\right\}.
\label{eq:par_plan}
\end{equation}
The branches start together, but actions inside the fetcher branch remain
ordered.  Grasping requires arrival at $g_p$, and carrying requires successful
grasp and lift.  Fetcher manipulation does not wait for guider arrival.  The
schedules thus satisfy the same evaluated destination-visit and delivery
objectives through different role assignments and temporal structures.

The scheduling hypothesis is that assigning the unloaded visit to the guider
allows it to overlap with object retrieval and can shorten a successfully
completed mission.  Let $T_{\mathrm F}^{\mathrm{visit}}$ and
$T_{\mathrm F}^{\mathrm{return}}$ denote the fetcher's unloaded trip to $g_f$
and return to $g_p$.  Let $T_{\mathrm F}^{\mathrm{move}}$ denote its direct move
from the initial position to $g_p$.  The remaining fetcher durations are grasp,
lift, and pre-transport retention, $T_{\mathrm F}^{\mathrm{grasp}}$, and loaded transport,
$T_{\mathrm F}^{\mathrm{carry}}$.  With guider navigation time
$T_{\mathrm G}^{\mathrm{nav}}$, the schedule structures give
\begin{align}
T_{\mathrm{seq}}={}&T_{\mathrm F}^{\mathrm{visit}}
 +T_{\mathrm F}^{\mathrm{return}}+T_{\mathrm F}^{\mathrm{grasp}}
 +T_{\mathrm F}^{\mathrm{carry}},
\label{eq:seq_time}\\
T_{\mathrm{par}}={}&\max\!\left(
T_{\mathrm G}^{\mathrm{nav}},
T_{\mathrm F}^{\mathrm{move}}+T_{\mathrm F}^{\mathrm{grasp}}
 +T_{\mathrm F}^{\mathrm{carry}}\right).
\label{eq:par_time}
\end{align}
The durations are measured during execution and need not match across robots or
scenes.  The comparison therefore captures the implemented combination of
heterogeneous role assignment and action overlap in the same mission.

Mission success requires state verification of every terminal objective, not
only a valid plan or completed controller command. Let $b_{\mathrm{visit}}$,
$b_{\mathrm{pickup}}$, $b_{\mathrm{grasp}}$, $b_{\mathrm{lift}}$,
$b_{\mathrm{retain}}$, and $b_{\mathrm{delivery}}$ be Boolean variables
computed from controller or simulator state. Operational success is
\begin{equation}
y_{\mathrm{op}} =
b_{\mathrm{visit}}
\land b_{\mathrm{pickup}}
\land b_{\mathrm{grasp}}
\land b_{\mathrm{lift}}
\land b_{\mathrm{retain}}
\land b_{\mathrm{delivery}}.
\label{eq:operational_success}
\end{equation}
The visit and pickup variables record arrival at $g_f$ and $g_p$. The remaining
variables record grasp, lift, pre-transport retention, and assisted delivery
under the state checks defined in Sec.~\ref{sec:system}. An action completes at the first simulator timestamp
when its terminal predicate becomes true.  Parallel makespan is the later of
guider arrival and fetcher delivery, while sequential makespan ends at delivery
after all preceding actions have completed.  Route length remains a trajectory
measurement and is not an additional success condition after goal arrival.  A
timeout or failed manipulation gate produces a failure record, never an
artificially short completion time.  Speed is therefore evaluated only after
both the destination visit and assisted object delivery have been verified.
\section{System Architecture}
\label{sec:system}

\method{} turns one user instruction into a verified guide-and-fetch mission by
validating schedule-conditioned action records, compiling their dependencies,
executing the robot branches under a shared clock, and recording completion
from robot and object states. Figure~\ref{fig:overview} summarizes this path
from language input to embodied outcome.

\subsection{Plan Instantiation and Validation}

Plan instantiation yields a validated four-action schema, not merely a raw LLM response.
The execution schedule is selected as an experimental condition before
prompting; the LLM does not choose it. For each condition, the prompt lists the
available robots, four permitted skills, registered targets, selected schedule,
and required JavaScript Object Notation (JSON) format. Each proposed action contains a step, robot, skill,
target, and parameter record. The JSON \texttt{step} field supplies the nominal
dependency-level index $k_j$. The parser extracts one JSON object, normalizes
robot, skill, and target aliases, and resolves the targets against the scene
registry.  The validator then checks
that all four expected actions are present, each assignee provides the requested
skill, and the carry action retains the registered object.  A parallel plan
must contain guider navigation and fetcher motion at step 1, followed by fetcher
grasp and carry.  A sequential plan must contain four ordered fetcher actions.
Extra or missing actions, incorrect targets, or incompatible robot assignments
cause rejection before simulation.  Rejected responses and validation messages
are stored, and any new response undergoes the same checks.  Scripted plans pass
through the same parser and validator as online LLM outputs.

The compiler uses the validated step indices to construct nominal dependency
levels and assigns each action a stable identifier stored in the run record. Parallel
plans contain three levels with two, one, and one actions, while sequential
plans contain four single-action levels. These levels describe the dependency
structure and do not impose global synchronization barriers. The two initial
parallel actions start together. Fetcher grasp and carry depend only on the
preceding fetcher events, so guider arrival does not delay manipulation.

\subsection{Schedule Execution}
\label{sec:executor}

The executor advances all active robots under one simulation clock while each
branch progresses from its own state events.  Sequential execution instantiates
only the fetcher, which visits the final station, returns to pickup, grasps and
lifts the object, and then delivers it.  Parallel execution instantiates both
robots in the same InternUtopia task.  Guider navigation and fetcher motion to
pickup begin at the same simulation step.  At every subsequent step, the
executor updates the active controllers and evaluates their terminal
predicates.  A satisfied predicate is timestamped immediately and releases the
next action on that branch.  The guider may reach the final station and hold
position while retrieval continues, and the fetcher may begin manipulation
while the guider is still moving.  No fixed waiting interval is inserted.  The
parallel mission completes only after both guider arrival and object delivery.

Failure propagation follows the same dependencies, so progress on one branch
cannot conceal an unmet objective on the other. A pickup timeout prevents
grasp and carry from starting. Missing bilateral contact or retained lift
prevents assisted transport. A missing guider arrival leaves the visit
objective unsatisfied even if the fetcher reaches the final station. The
executor records the branch progress and first failed gate before ending the
episode. Both schedules use the same fetcher controller, scene
configuration, sampled perturbation, and completion thresholds.
\subsection{Robot Skills}

In the parallel schedule, the guider performs the unloaded destination visit,
with success defined by goal attainment under the implemented registered-route
controller. It uses a recurrent policy trained with Proximal Policy Optimization
(PPO)~\cite{schulman2017ppo} from 180 planar light detection and ranging
(lidar)
measurements and an 11-element state vector that encodes goal geometry, base
velocity, body orientation, and the previous action. During the shared mission,
the policy contributes local planar commands and the visible gait, while a
waypoint progress shield advances the root through kinematic transport along
the registered route. Collision response for the transported root is disabled.
The navigation condition is satisfied when
the guider reaches within $0.20$~m of the final waypoint, so this event records
destination arrival and does not establish dynamic collision-aware locomotion.

The fetcher provides object acquisition and assisted delivery, with contact and lift
verified before assisted transport begins.  It combines a locomotion policy
associated with Autonomous Large-Object Rearrangement with a Legged Manipulator
(ALORE)~\cite{bi2026alore} with joint-space arm staging and world-frame inverse
kinematics. The ALORE-derived controller supplies leg commands and visible
gait, while registered-route transport advances the base to pickup and delivery.
The base stops before the arm aligns the gripper with the registered object
handle. Grasp success requires contact on both sides of the object. Lift success
requires a peak object rise of at least $30$~mm. Retention is evaluated over a
1.0~s hold at 200~Hz and requires at least $20$~mm retained rise, bilateral
contact in at least 80\% of hold frames and in the final frame, at least 5~mm
separation between the target and support collision bounds, and at most
$17^\circ$ object tilt. The distance from the
registered grasp reference to the end-effector frame origin must also remain
within $0.25$~m. This empirically configured grasp-reference consistency bound
accommodates the fixed offset from the frame origin to the gripper contact
region and rejects gross object separation. It is neither contact evidence nor
a fingertip-clearance measure.
Object pose synchronization with the end effector starts only after these
pre-transport checks succeed. The initial contact, lift, and retention are
measured from simulated state, while subsequent transport is assisted along a
registered delivery route. Once acquisition begins, both schedules use the same
manipulation sequence, thresholds, and assisted-delivery controller.

\subsection{Completion Checks and Run Records}

Completion checks provide one evidence layer for both schedules and connect the
validated plan to the final mission outcome.  The unloaded visit is satisfied
when its assigned robot reaches the final station without the object.  Pickup
arrival is recorded when the fetcher completes its registered route to $g_p$.
Grasp requires bilateral contact, lift requires the measured object rise, and
retention requires the hold gate defined above. Delivery requires the fetcher
and synchronized object state to reach the final station after that gate has
passed. These measurements determine the variables in
Eq.~\eqref{eq:operational_success}. A failed check ends the episode and prevents
dependent actions from running. Route length and waypoint progress are
stored for analysis but do not add a second success condition after goal
arrival.  Each run record stores the scene, seed, plan provenance, schedule,
normalized plan, validation result, controller configuration, action
transitions, terminal conditions, timings, and object lift measurements.  An
online LLM record also retains the raw response, model identifier, request
latency, token counts, and rejected attempts. Common identifiers for scene, seed,
plan provenance, and schedule link these fields to one episode, allowing a failure to be
traced to planning, controller execution, or a state-based completion check.
\section{Experiments}
\label{sec:experiments}

The experiments test normalized-plan agreement and validator rejection, then compare
the implemented sequential and role-reassigned parallel schedules under common
state-based completion conditions.

\subsection{Implementation Details}

\begin{figure}[t]
\centering
\includegraphics[width=\textwidth]{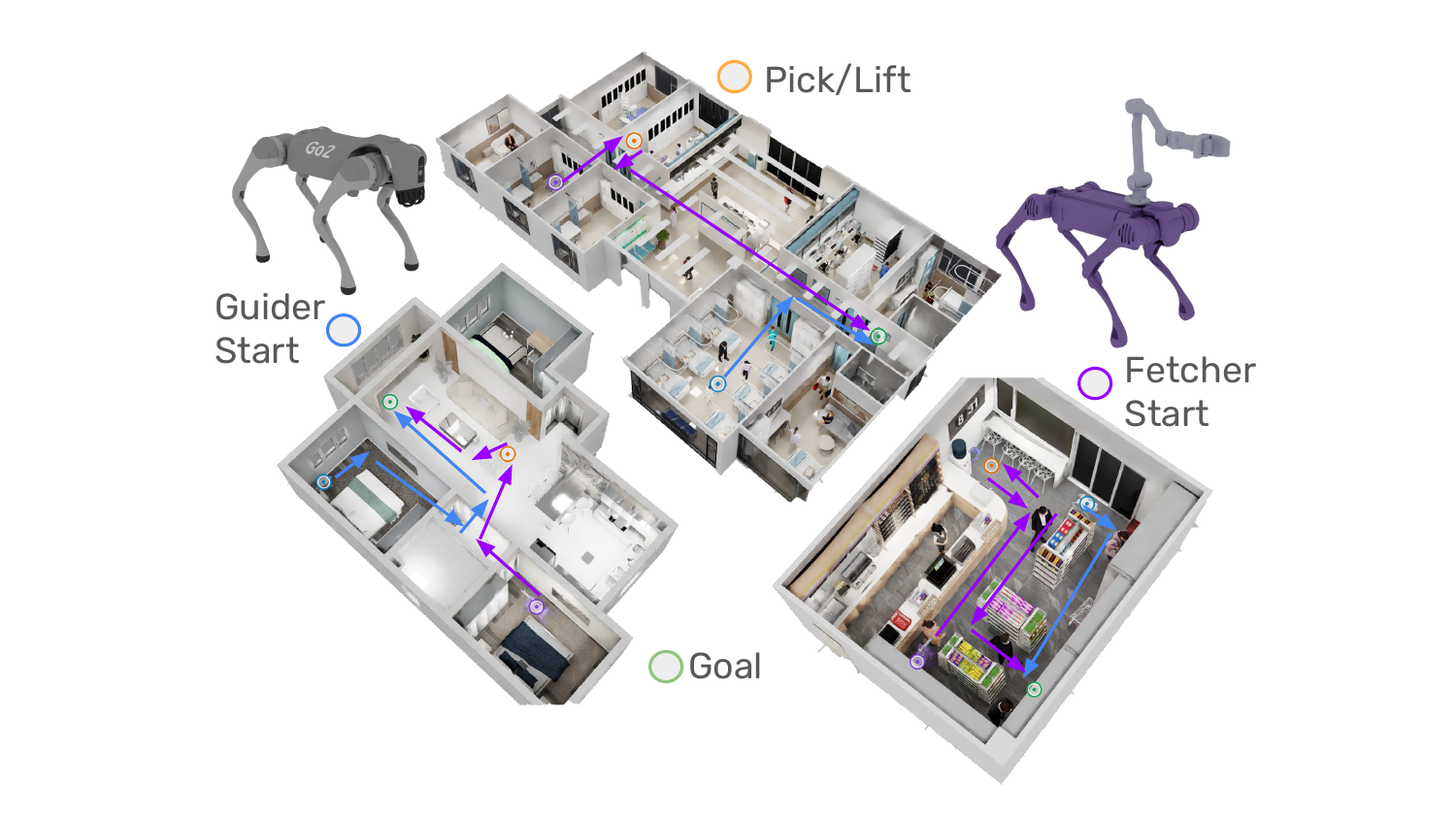}
\caption{Registered layouts in Hospital (top), Home (lower left), and
Supermarket (lower right). Blue arrows show the guider route to the shared
goal, and purple arrows show the fetcher route through pickup to that goal.
Parallel execution uses both routes. Sequential execution uses the same
locations but assigns both objectives to the fetcher.}
\label{fig:scene_routes}
\end{figure}

The implementation fixes scene geometry and routes before the plan-provenance
and schedule comparisons.  We use the official Home, Supermarket, and Hospital
GRScenes environments. Figure~\ref{fig:scene_routes} shows the registered robot
starts, pickup location, shared goal, and role-specific routes in each scene.
The object pose at pickup and all route waypoints are also registered. The
registered paths satisfy the scene-specific clearance requirements and are
shared by both plan-provenance conditions. The longer Hospital delivery route is retained
as part of the original scene layout.

The target is a runtime-spawned $60\times74\times85$~mm handled calibration
object weighing 5.00--5.25~g. It is not a filled beverage container; coffee
delivery motivates the task but does not describe the evaluated payload.

All experimental conditions use the same simulator and controller settings.
The simulator advances at 200~Hz, and both robots in the parallel condition
share one physics clock.  Background scene physics and collision response for
route-based base transport are disabled, while robot articulation,
target-object dynamics, and gripper contact remain active.  The manipulation
controller and its success thresholds are fixed across conditions.  The object
follows the end effector during transport only after the grasp, lift, and hold
conditions have passed.  Because collision response is inactive for the
registered-route transport, collision statistics are not reported.

\subsection{Experimental Design}

Scripted and online plan-provenance conditions test normalized-plan agreement, whereas the schedule
comparison measures sequential and parallel execution as defined in
Eqs.~\eqref{eq:seq_plan} and~\eqref{eq:par_plan}.  The online language interface uses a
locally served four-billion-parameter Qwen3.5 model~\cite{qwen2026qwen35}. The
180 artifacts record the model identifier \path{frob/qwen3.5-instruct:4b}
and the same local endpoint. All four conditions use the same normalization,
validation, controller checkpoints, state checks, and timeout.  Every Qwen3.5
run makes a new request without fallback or recorded replay.  Four semantically
equivalent instruction templates per schedule are rotated across seeds to vary
the wording while preserving the requested subgoals.  We use 30 seeds in each
of the three scenes.  Each seed produces all four conditions, giving 90 matched
sets and 360 separate simulator executions.

Every scene and seed is matched across the four conditions, so a comparison is
not confounded by a different initialization.  Seeded perturbations vary robot
starts by no more than 10~mm, object translation by 1.5~mm, object yaw by
$0.5^\circ$, and object mass between 5.00 and 5.25~g.  Contact friction is also
varied within a narrow calibrated range.  The exact perturbation record is
generated once per scene and seed and copied into all four launch
configurations. Plan-provenance comparisons match scene, seed, and schedule. Schedule
comparisons match scene, seed, and plan provenance. All 90 sets contain the
expected four records.  Failed episodes remain in the success-rate denominator
and are never replaced by additional runs. For the validator test, we start
from one valid control per schedule and construct single-edit mutations that
alter the action inventory (10), schedule structure (8), registered target (8),
capability assignment (4), or carried-object identity (2), while leaving the
remaining fields unchanged.

\subsection{Evaluation Metrics}

Plan validity and operational success are reported separately, so a correct
action schema is not mistaken for a completed mission. A plan is valid when it
contains one parseable JSON object, the expected four actions, valid robot
assignments, and a structure admissible under the selected schedule. Online runs must also retain the raw
response and request metadata and show that neither fallback nor replay was
used.  Operational success follows Eq.~\eqref{eq:operational_success} and
requires goal arrival, pickup arrival, grasp, lift, hold, and delivery under the
checks in Sec.~\ref{sec:system}.  Path length and waypoint progress are
trajectory measurements, not additional success criteria after goal arrival.

Makespan is compared only among matched runs completed by both schedules, which
prevents early failure from appearing as a timing improvement.  It is measured
from the end of simulator stabilization until the last required terminal event.
Relative time saving is $1-T_{\mathrm{par}}/T_{\mathrm{seq}}$.  Success rates
are reported with Wilson 95\% confidence intervals.  Paired mean differences
use a deterministic 20,000-sample percentile bootstrap interval and a
50,000-draw sign-flip test, with scripted and online results analyzed
separately.  Trajectory logs also separate the fetcher's unloaded visit, pickup
approach, and loaded delivery from the guider route.  Total distance sums the
paths of all robots active in a condition and remains descriptive because two
robots can travel farther in total while completing their routes concurrently.
\section{Results}
\label{sec:results}

\subsection{Plan Validity and Operational Success}

\textbf{Plan validity does not imply mission success.}
All 180 online Qwen3.5 responses retain the raw output and prompt hash, parse as
one JSON object, pass all checks on the first request, and normalize to the
scripted reference. The two plan-provenance conditions are separate simulator
executions, but after normalization they provide identical controller inputs
and reproduce identical outcomes. Table~\ref{tab:main_results} therefore reports
success once per scene--seed case within each schedule; execution timing starts
after validation.

\begin{table}[!htbp]
\centering\small
\caption{Normalized-plan agreement and operational success. Success is reported
once per scene--seed case within each schedule because both plan-provenance
conditions yield identical controller inputs and outcomes.}
\label{tab:main_results}
\begin{tabular}{@{}l@{\hspace{1.6em}}c@{\hspace{1.6em}}l@{}}
\toprule
Evaluation & Scope & Result \\
\midrule
Normalized-plan agreement & 180 pairs & $180/180$ matched \\
Sequential success & 90 cases & $72/90$ ($80.0\%$; 95\% CI $[70.6, 87.0]$) \\
Parallel success & 90 cases & $71/90$ ($78.9\%$; 95\% CI $[69.4, 86.0]$) \\
\bottomrule
\end{tabular}
\end{table}

\textbf{Controlled invalid plans exercise validator rejection.}
Both valid schedule controls are accepted and all 32 single-edit mutations are
rejected. This simulator-free test demonstrates rejection on
the controlled set, not open-ended understanding or repair.

\subsection{Paired Makespan}

\textbf{The implemented parallel protocol is faster in every common-success
case.}
Across 56 matched cases, mean makespan decreases from 164.78 to 96.79~s
($1.70\times$; $41.3\%$). The paired reduction is 67.99~s, with bootstrap 95\%
interval $[54.05,82.04]$~s and sign-flip $p<0.001$. Because schedules also
differ in assignment and, in Hospital, route length, this is a system-level
result rather than an isolated concurrency effect (Table~\ref{tab:scene_results}).

\begin{table}[!htbp]
\centering\small
\caption{Common-success timing and total robot distance. Seq. and Par. denote
sequential and parallel schedules; ratio is Seq. makespan divided by Par.
makespan, and distance sums all active robots.}
\label{tab:scene_results}
\resizebox{\textwidth}{!}{%
\begin{tabular}{lcccccccc}
\toprule
Scene & Seq. success & Par. success & Pairs & Seq. [s] & Par. [s] & Ratio & Seq. [m] & Par. [m] \\
\midrule
Home & $20/30$ & $21/30$ & 13 & 114.13 & 69.04 & $1.65\times$ & 22.05 & 24.95 \\
Supermarket & $26/30$ & $25/30$ & 22 & 84.02 & 67.94 & $1.24\times$ & 14.86 & 16.76 \\
Hospital & $26/30$ & $25/30$ & 21 & 280.74 & 144.20 & $1.95\times$ & 68.25 & 40.60 \\
\bottomrule
\end{tabular}}
\end{table}

\textbf{The source of the timing benefit depends on scene geometry.}
In Home and Supermarket, time falls by 39.5\% and 19.1\% although total distance
rises from 22.05 to 24.95~m and 14.86 to 16.76~m, respectively, supporting an
overlap benefit. In Hospital, role reassignment shortens total travel from 68.25
to 40.60~m while time falls by 48.6\%. The aggregate gain therefore combines
assignment, overlap, and scene geometry.

\subsection{Manipulation Outcomes}

\textbf{Schedule outcomes are discordant for 31 of 90 matched seeds.}
Of the 90 pairs, 56 succeed under both schedules, 16 only sequentially, 15 only
in parallel, and 3 under neither.

\begin{figure}[!t]
\centering
\includegraphics[width=\textwidth]{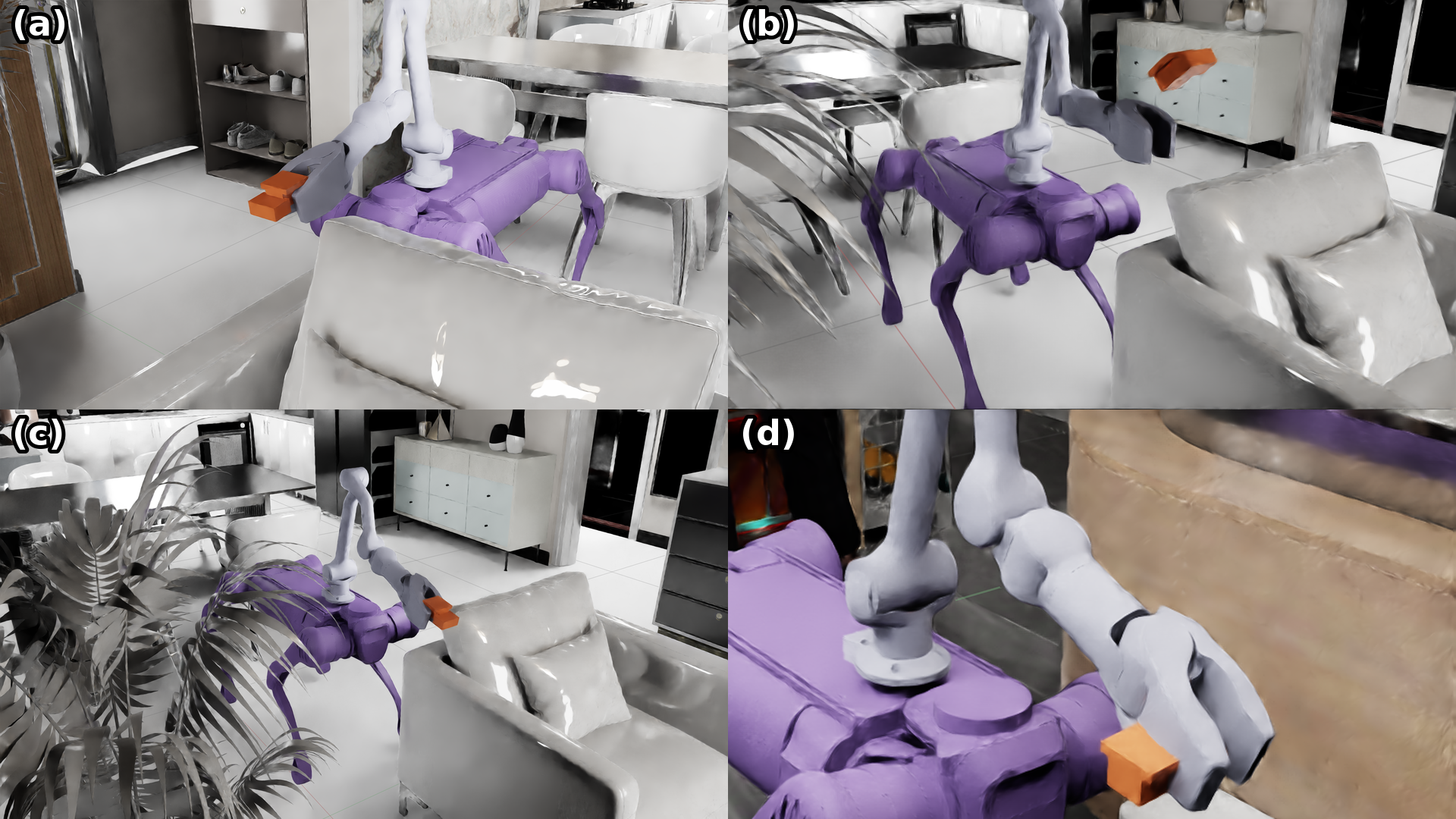}
\caption{Representative simulator captures of the manipulation outcomes:
(a) verified retention, (b) grasp-reference consistency failure (19),
(c) excessive hold tilt (15), and (d) loss of final bilateral contact
(3; 2 also exceed the tilt limit).}
\label{fig:failure_cases}
\end{figure}

\textbf{All operational failures arise during manipulation.}
We count the 37 distinct scene--schedule--seed failures once: 19 exceed the
0.25~m grasp-reference consistency bound during retention, 15 retain final bilateral
contact but exceed $17^\circ$ tilt, and three lose final bilateral contact (two
also exceed tilt). These comprise 18 sequential and 19 parallel cases. No
simulator episode fails validation or pickup arrival, and every parallel guider
reaches its goal; the success difference therefore comes from fetcher
manipulation (Figure~\ref{fig:failure_cases}).

\textbf{Successful episodes verify grasp and lift before assisted transport.}
The 143 successful scripted episodes comprise 41 Home, 51
Supermarket, and 51 Hospital runs.  All pass the bilateral-contact, lift, and
hold checks before post-lift object synchronization.  Mean peak and retained
object rise are 69.89 and 62.05~mm in Home, 66.48 and 61.89~mm in Supermarket,
and 69.43 and 63.05~mm in Hospital.  Corresponding online runs produce the same
measurements under the normalized plans, so Table~\ref{tab:lift_results} reports
one plan-provenance condition.

\begin{table}[!htbp]
\centering\small
\caption{Object rise in successful scripted episodes, measured before post-lift
object synchronization.}
\label{tab:lift_results}
\begin{tabular}{lccc}
\toprule
Scene & Successful episodes & Peak rise [mm] & Retained rise [mm] \\
\midrule
Home & 41 & 69.89 & 62.05 \\
Supermarket & 51 & 66.48 & 61.89 \\
Hospital & 51 & 69.43 & 63.05 \\
\bottomrule
\end{tabular}
\end{table}
\section{Discussion}
\label{sec:discussion}

\textbf{The language interface is consistent within the tested schema.}
All 180 responses match the scripted references, but the schedule is supplied
to the prompt rather than chosen by the LLM. This establishes schema recovery,
not open-ended decomposition; the mutation test covers controlled invalid
inputs, not generalization.

\textbf{The timing gain has two sources.} In Home and Supermarket, the parallel team
travels farther in total yet finishes sooner, showing the effect of temporal
overlap. In Hospital, role reassignment also removes the fetcher's long unloaded
visit and return. The $41.3\%$ reduction therefore reflects the implemented
role assignment, action overlap, and scene geometry, not faster controllers.

\textbf{Manipulation remains the bottleneck.}
Success includes all 90 cases per schedule, while makespan uses the 56 common
successes. Every failure occurs during pre-transport grasp or retention; every
parallel guider reaches its goal. Object handling therefore limits completion.

The 31 discordant seed pairs share perturbations but follow different fetcher
routes and pre-acquisition histories; this study does not isolate the causal
state difference.
\section{Limitations and Future Work}
\label{sec:limitations_future_work}

\textbf{Language generalization.}
Evidence for the language interface is limited to four paraphrase templates per schedule and a fixed action inventory. 
The mutation test covers deterministic rejection of controlled plan violations,
not LLM behavior on unseen schemas, infeasible or conflicting requests, or plan
repair. Future experiments should introduce held-out instructions, larger action
schemas, and state feedback that requires replanning.

\textbf{Manipulation scope.}
The manipulation evidence covers one lightweight handled object under bounded pose, mass, and friction perturbations. Physical checks end before assisted transport begins. 
Future work should test varied objects, require unaided retention throughout loaded motion, and verify delivery until object release at the destination.

\textbf{Deployment realism and human guidance.}
The deployment study is limited to three registered scenes, controlled routes, disabled background physics, and kinematic root transport. 
Destination arrival is only a proxy for guidance because human following and escort safety are not modeled. 
Future work should add dynamic base control, changing obstacles, and unseen layouts before user studies examine task time, uncertainty, workload, and perceived risk with blind and low-vision participants.
\section{Conclusion}
\label{sec:conclusion}

This work introduced \method{}, a framework for coordinating heterogeneous guider and fetcher robots from a single natural-language request. 
The framework instantiates a schedule-conditioned action schema, checks each assignment against robot capabilities, and coordinates sequential or parallel execution through state-based events. 
It keeps plan validity, execution progress, and verified completion distinct. 
In the motivating scenario, one robot visits the destination while another performs state-verified object acquisition and assisted delivery. 
\method{} thus integrates role specialization and concurrent action without treating a valid plan as evidence of mission success.

\section{Acknowledgment}

This work was supported in part by the Ministry of Science, Research and the Arts of Baden-W\"urttemberg (MWK) through the Cooperative Graduate School Accessibility through AI-based Assistive Technology (KATE) under Grant BW6-03, in part by funding from the pilot program Core-Informatics of the Helmholtz Association (HGF), in part by Karlsruhe House of Young Scientists (KHYS), and in part by the Helmholtz Association Initiative and Networking Fund on the HAICORE@KIT and HOREKA@KIT partition. 
This work is supported in part by the Deutsche Forschungsgemeinschaft (DFG, German Research Foundation) - SFB 1574 - 471687386.
This work was also supported in part by the National Natural Science Foundation of China (Grant No. 62473139), in part by the Hunan Provincial Research and Development Project (Grant No. 2025QK3019), and in part by the State Key Laboratory of Autonomous Intelligent Unmanned Systems (the opening project number ZZKF2025-2-10).

\bibliographystyle{splncs04}
\bibliography{references}

\end{document}